\documentclass[sn-basic,iicol]{sn-jnl}

\usepackage{graphicx}%
\usepackage{multirow}%
\usepackage{amsmath,amssymb,amsfonts}%
\usepackage{amsthm}%
\usepackage{mathrsfs}%
\usepackage[title]{appendix}%
\usepackage{xcolor}%
\usepackage{textcomp}%
\usepackage{manyfoot}%
\usepackage{booktabs}%
\usepackage{algorithm}%
\usepackage{algorithmicx}%
\usepackage{algpseudocode}%
\usepackage{listings}%

\usepackage{natbib}
\usepackage{url}
\usepackage{mathtools}
\usepackage{balance}
\usepackage{hhline}
\usepackage{makecell}
\usepackage{epsfig}
\usepackage{subcaption}
\usepackage{multirow}
\usepackage{multicol}
\usepackage{makecell}
\usepackage{xcolor, soul}
\usepackage{array}
\usepackage{hhline}

\usepackage[accsupp]{axessibility}

\begin{document}

\title[Article Title]{Few-Shot Referring Video Single- and Multi-Object Segmentation via Cross-Modal Affinity with Instance Sequence Matching}


\author[1]{\fnm{Heng Liu} } \email{hengliu@ahut.edu.cn}

\author[1]{\fnm{Guanghui Li} }\email{guanghui.li1998@gmail.com}

\author[2]{\fnm{Mingqi Gao} }\email{mingqi.gao@outlook.com}

\author[2]{\fnm{Xiantong Zhen} }\email{zhenxt@gmail.com}

\author[2]{\fnm{Feng Zheng} }\email{f.zheng@ieee.org}

\author[3,*]{\fnm{Yang Wang}}\email{yangwang@hfut.edu.cn}

\affil[1]{\orgdiv{School of Computer Science and Technology}, \orgname{Anhui University of Technology}, \orgaddress{\street{Maxiang Road}, \city{Ma'anshan}, \postcode{243032}, \country{China}}}

\affil[2]{\orgdiv{Department of Computer Science and Engineering}, \orgname{Southern University of Science and Technology}, \orgaddress{\street{Xueyuan Avenue}, \city{Shenzhen}, \postcode{518055}, \country{China}}}

\affil[3]{\orgdiv{School of Computer Science and Information Engineering}, \orgname{Hefei University of Technology}, \orgaddress{\street{Feicui Road}, \city{Hefei}, \postcode{230601}, \country{China}}}
\affil[*]{Corresponding Author}
\abstract{
    Referring Video Object Segmentation (RVOS) aims to segment specific objects in videos based on the provided natural language descriptions. As a new supervised visual learning task, achieving RVOS for a given scene requires a substantial amount of annotated data. However, only minimal annotations are usually available for new scenes in realistic scenarios. Another practical problem is that, apart from a single object, multiple objects of the same category coexist in the same scene. Both of these issues may significantly reduce the performance of existing RVOS methods in handling real-world applications. In this paper, we propose a simple yet effective model to address these issues by incorporating a newly designed cross-modal affinity (CMA) module based on a Transformer architecture. The CMA module facilitates the establishment of multi-modal affinity over a limited number of samples, allowing the rapid acquisition of new semantic information while fostering the model’s adaptability to diverse scenarios. Furthermore, we extend our FS-RVOS approach to multiple objects through a new instance sequence matching module over CMA, which filters out all object trajectories with similarity to language features that exceed a matching threshold, thereby achieving few-shot referring multi-object segmentation (FS-RVMOS). To foster research in this field, we establish a new dataset based on currently available datasets, which covers many scenarios in terms of single-object and multi-object data, hence effectively simulating real-world scenes. Extensive experiments and comparative analyses underscore the exceptional performance of our proposed FS-RVOS and FS-RVMOS methods. Our method consistently outperforms existing related approaches through practical performance evaluations and robustness studies, achieving optimal performance on metrics across diverse benchmark tests.    
}
\maketitle
\setcounter{page}{2}
\section{Introduction}\label{sec:introduction}
Referring Video Object Segmentation (RVOS) aims to segment the target objects described in natural language within video content, which supports the wide applications in real-world scenarios, including video editing (\cite{fu2022m3l}) and human-computer interaction, thus garnering significant attention from the research community. Compared to traditional semi-supervised video object segmentation (VOS) methods (\cite{gao2023deep}), RVOS introduces significant challenges by requiring the integration of both visual and linguistic cues, while also lacking ground-truth masks for the initial video frame. However, detailed annotated masks and their corresponding critical language descriptions, which are essential for real-world RVOS tasks, remain relatively scarce. Obtaining high-quality labelled data necessitates a high cost, as the researchers have to meticulously annotate each frame in the video with details to offer a referring expression for the segmentation object. At the same time, due to the widespread popularity of movies, such as YouTube videos, TikTok streams, etc., the volume of video data across various domains has witnessed explosive growth, making the demand for diverse video segmentation extremely challenging.

To handle broad-ranging video data, existing RVOS methods (\cite{botach2022end,wu2022language}) rely on massive and diverse labels for training. However, these methods are fundamentally limited by fixed and restricted training classes, which hinders their ability to adapt to the dynamic and diverse real-world data landscape. On the other hand, fine-tuning existing RVOS methods on a limited number of samples to accommodate real-world scenes often falls short of producing high-quality results since a small amount of labeled data may be insufficient to facilitate the model's understanding of new semantic information.
In addition, few works currently can effectively handle the referring video segmentation for multiple objects of the same category. The notorious situations, such as occlusion, make achieving accurate reference segmentation of objects with limited samples more challenging.
Therefore, it is urgent to find a way to make RVOS methods more applicable to various scenes in the real world at a lower cost.

In fact, inspired by few-shot semantic segmentation (FSS) (\cite{zhou2024unlocking,liu2022dabdetr,tong2024dynamic,liu2024ntrenet++}), few-shot learning enables models to generalize effectiveness from a minimal amount of labeled data, reducing the need for large datasets and accelerating adaptation to new scenes, which sheds light on the RVOS task.
\begin{figure}[bt]
  \centering
  \includegraphics[width=0.86\linewidth]{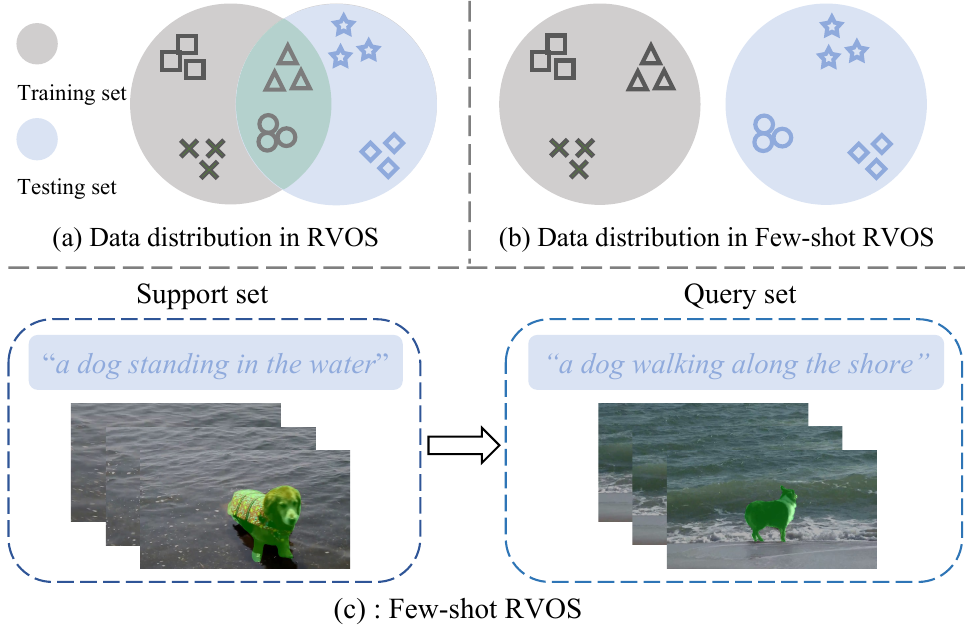}
   \caption{Comparison of Few-shot RVOS and RVOS.
(a) The training and testing sets overlap in the RVOS.
(b) Disjoint training and testing sets in the Few-shot RVOS. 
Different shapes represent different classes.
(c) Few-shot RVOS segments the referred object of the same class as the support set in the video.}
   \label{fig:figure1}
\vspace{-3ex}
\end{figure}

Thus, we formulate the above problem as Few-Shot Referring Video Object Segmentation (FS-RVOS), to delineate the FS-RVOS scenario and highlight its distinctions from conventional RVOS in Figure \ref{fig:figure1}. Unlike RVOS, in FS-RVOS, the categories in the training and testing sets are mutually exclusive. With access to a limited number of supporting video clips along with their corresponding language descriptions and object masks, FS-RVOS aims to segment videos within the query set, as illustrated in Figure \ref{fig:figure1}(c).\par

For FS-RVOS, there are currently two routes based on prototype and attention mechanisms, both of which harness the vision-language information within the support set from different aspects.
Prototype-based methods (\cite{nguyen2019feature,yang2020prototype}) compress features from different classes to generate prototypes, which, however, disregard spatial structures and suffer from noise, leading to varying degrees of information loss.
Another route (\cite{zhang2019pyramid,zhang2021few,zhang2022catrans}) employs attention mechanisms to encode foreground pixels from support features and integrate them with query features. Although these methods have shown high-quality results in both the image and video domains, their application in vision-language tasks remains relatively unexplored.

To this end, we develop a cross-modal affinity (CMA) module (\cite{li2023learning}) to capture multi-modal relationships from a limited number of samples and extract new semantic information for diverse data. Inspired by RVOS, we take advantage of the complementary strengths of visual and linguistic modalities to enhance affinity mining across modalities, thus achieving more robust segmentation in scenarios with few annotated video frames. Throughout only a handful of annotated samples (comprising language expressions and referred object masks), we employ a hierarchical fusion strategy to cross-attend visual and textual features for robust feature representations tailored to specific categories, further enabling the model to process large quantities of data within the same category efficiently. Due to CMA, multi-modal features of the support and query sets are fused independently and then aggregated through their interrelationships to alleviate the irrelevant features.

In real-world scenarios, it is common for multiple objects of the same category to appear simultaneously in a video, which causes prior single-object methods to struggle to perform satisfactorily in multi-object scenarios. To accommodate scenarios involving the simultaneous presence of multiple objects, we need to extend the research on few-shot referring video single-object segmentation and achieve accurate segmentation of multiple objects in videos based on natural language descriptions.
Therefore, upon our previous research (\cite{li2023learning}), this work develops a new instance sequence matching (ISM) module over CMA to solve the problem of multi-object segmentation in videos with limited samples. Specifically, we use a linear projection network to generate the referring matching scores after the Transformer's decoding. We compare the score with the threshold to determine which instance trajectories should be retained and which can be discarded. In addition, different weights are used to balance the trajectories of multi-objects, leveraging complementary information among them to enhance the segmentation integrity of multi-objects. Thus, our approach cannot only select the best matching trajectory but also pick out all object trajectories with similarity to referring text features exceeding a matching threshold $\sigma$. Please take a look at Figure \ref{fig:figure35} in Section 3.5 for detailed reference. The proposed ISM module performs, after CMA, to handle single-object segmentation tasks flexibly and cope with multi-objective segmentation for FS-RVMOS.

Currently, since there is no dataset for the task of few-shot RVOS, we build up a new FS-RVOS benchmark based on Ref-YouTube-VOS (\cite{seo2020urvos}), named Mini-Ref-YouTube-VOS. 
To measure the model's generalization ability, we also construct a dataset which is different from natural scenes based on a synthetic dataset SAIL-VOS (\cite{hu2019sail}), named Mini-Ref-SAIL-VOS. 
The newly constructed two benchmark datasets cover a wide range with a balanced number of high-quality videos in each category.
Based on our previous work (\cite{li2023learning}), to further simulate multi-object scenes in the real world, we construct a new FS-RVMOS dataset, named Mini-MeViS following MeViS (\cite{ding2023mevis}), which focuses on the existence of multiple similar objects.

The major contributions of this work are summarized as follows.
\begin{itemize}
\item For limited samples in the real world, we propose a \textbf{C}ross-\textbf{M}odal \textbf{A}ffinity (CMA) module combined with a new \textbf{I}nstance \textbf{S}equence \textbf{M}atching (ISM) structure to encode the affinity of multiple modalities, allowing flexible referring video segmentation of both single and multiple objects.
\item We explore a novel few-shot RVOS problem, which learns new semantic information over limited samples to adapt to real-world scenarios with few annotations and dynamic and unrestricted categories.
\item We build up the first FS-RVOS and FS-RVMOS benchmarks, where we conduct comprehensive comparisons with existing methods in diverse real scenarios, showing the superiority of the proposed model. 
\end{itemize}

\section{Related work}\label{sec:related_work}
    \subsection{Few-Shot Semantic Segmentation}
    Few-shot semantic segmentation, initially proposed by \cite{shaban2017one}, aims to learn to segment new image categories with only a few samples. Afterwards, early advances in this field stem from the adoption of metric learning techniques.
    Building on the PrototypicalNet framework, \cite{dong2018few} utilize metric learning and cosine similarity between pixels and prototypes for prediction. Then, to simplify the framework, \cite{wang2019panet} introduce the prototype alignment regularization and propose PANet.
    \cite{tian2020prior} propose the PFENet model, which leverages prior knowledge from pre-trained backbones to identify regions of interest and employs various designs of feature pyramid modules, utilizing the mappings from previous stages to enhance segmentation performance. \par

    However, the efficacy of existing few-shot segmentation methods heavily relies on the quality of prototypes derived from the support set. To address this, \cite{fan2022self} develop a self-support matching method with query features, alleviating intra-class appearance differences inherent in few-shot segmentation. Their strategy can capture consistent underlying features of query objects for feature matching.
    Similarly, \cite{Tian_2022_CVPR} propose a new context-aware prototype learning method that leverages prior knowledge from support samples and dynamically enriches contextual information through adaptive features, thus enhancing the robustness of prototype learning by integrating contextual cues.
    Drawing upon the concept of utilizing a base learner to identify confusing regions within query images and refining predictions through a meta-learner, \cite{Lang_2022_CVPR} pioneers a novel method, named BAM, to few-shot segmentation. In contrast to conventional methods focusing on feature extraction or visual correspondence, BAM emphasizes iterative improvement of predictions, thereby improving the segmentation results. 
    
    Recently, motivated by cross-domain style transfer,  \cite{su2024domain} propose to train a domain-rectifying adapter to convert the diverse characteristics of the target domain to the source domain and leverage the well-trained source domain segmentation model to process the features of the rectified target domain for accurate few-shot semantic segmentation. Recognizing that pre-trained vision transformers inherently capture semantic visual groupings, \cite{zhou2024unlocking} propose an innovative adaptation of "relationship descriptors" to address overfitting in few-shot semantic segmentation tasks. Aiming to generalized few-shot semantic segmentation (GFSS), \cite{tong2024dynamic} design a dynamic knowledge adapter (DKA), which enhances model adaptability via efficient parameter fine-tuning and mitigates training instability through sample relabeling mechanisms, with correcting prediction bias using probabilistic calibration layers. In addition, to address the problems of coarse granularity and obvious category bias in prior encoding features in previous approaches, \cite{wang2024rethinking} present using the visual text alignment capability of the Contrastive Language-Image Pre-training model (CLIP) instead of visual prior representation to capture more reliable guidance and enhance model generalization performance. While existing studies in few-shot semantic segmentation predominantly focus on target object mining, they often fail to address ambiguities in non-target regions such as background (BG) and distracting objects (DO). To resolve this, \cite{liu2022learning,liu2024ntrenet++} introduce two successive frameworks (NTREnet and NTREnet++) that systematically identify and suppress non-target interference. Their approach combines a BG mining loss for the supervision of the BG mining module and a prototypical-pixel contrastive learning mechanism to enhance target-DO differentiation.\par

    Compared to image-based few-shot segmentation, investigation on few-shot video object segmentation is relatively scarce and remains nascent. The initial works (\cite{siam2021weakly,chen2021delving}) primarily address this challenge through attention mechanisms. However, these methods overlook temporal cues. Subsequently, leveraging temporal transductive reasoning, \cite{siam2022temporal} applies reasoning mechanisms and has shown promising results in cross-domain scenarios. Recent work like VIMPT (\cite{liu2023multi}) proposes to utilize multi-level temporal guidance to handle the temporal correlation of video data, which decomposes query video information into clip-level, frame-level, and memory-level prototypes, so as to leverage multi-level structures to capture local and global temporal relationships, respectively. Moreover, \cite{tang2024holistic} introduces a prototype graph attention module and a bidirectional prototype attention module, named HPAN, to transfer informative knowledge from seen to unseen classes, which not only enhances the overall prototype representation but also achieves support for query semantic consistency and intra frame temporal consistency.
    
    To sum up, existing few-shot segmentation methods predominantly focus on a single modality - either image or video and do not address the segmentation problem under multi-modal conditions, such as those involving linguistic referring expressions.
    \par
        
    \subsection{Referring Video Object Segmentation} 
    \cite{gavrilyuk2018actor} define the RVOS task. They generate convolution dynamic filters from textual representations and convolve them with visual features of different resolutions to obtain segmentation masks. In an effort to overcome the constraints of conventional dynamic convolution, \cite{wang2020context} propose a context-modulated dynamic convolution operation tailored for RVOS. Here, the kernel is derived from language sentences and surrounding contextual features. However, their approach is primarily geared towards video actors and actions, limiting its applicability to a few object classes and action-oriented descriptions. On the other hand, weak-shot semantic segmentation (WSSS) (\cite{chen2022weak,zhou2021weak}) takes a broader approach by focusing on the overall scene in the image. It treats masks and pixel-level annotations as the support set, with image-level texts serving as the query set. However, in WSSS, the text is confined to single words or phrases denoting class names, directly mapped to labels for pixel-level classification.
    
    \cite{khoreva2018rvos} develop a two-stage approach for RVOS, starting with referring expressions grounding and leveraging the predicted bounding boxes to guide pixel-wise segmentation. Similarly, \cite{seo2020urvos} present URVOS, a framework that predicts the initial masks at each frame based on the language expression, and then employs the predicted masks from the preceding frames for RVOS through the use of a memorized attention module. \par
    
    Recent advances in RVOS have seen a shift towards employing cross-attention mechanisms to facilitate interaction between visual images and linguistic information. For example, LBDT (\cite{ding2022language}) utilizes language as an intermediary bridge, connecting temporal and spatial information, and incorporates cross-modal attention operations to aggregate language-related motion and appearance cues. Similarly, MMVT (\cite{zhao2022modeling}) calculates optical flow between frames, integrating it as motion information with text and visual features. However, these frame-based spatial granularity multi-modal fusion methods have their limitations and often result in mismatches between visual and linguistic information. To address this challenge, \cite{wu2022multi} has explored novel multilevel representation learning methods and introduced dynamic semantic alignment techniques to adaptively fuse the two modalities, enhancing the consistency and effectiveness of the fusion process. \par

   Transformer (\cite{vaswani2017transformer}) has been widely applied and achieved great success in many computer vision tasks, such as object detection and tracking (\cite{carion2020detr,zhu2020deformable, wu2023referring, zhang2024bootstrapping}), image segmentation (\cite{SETR,cheng2021maskformer}), and image generation (\cite{liu2024structure}). Since DETR (\cite{carion2020detr}) introduces a new query-based paradigm, the latest works (\cite{botach2022end, wu2022language, li2023robust}) prefer to apply the DETR-like framework to the RVOS task.
    Specifically, they utilize Transformer structures to interact visual images with linguistic data and thereby are able to attain SOTA performance in accuracy and efficiency.
    However, these DETR structure-based works have their object queries at the frame level, independently responsible for searching for objects within each frame, failing to leverage the temporal information of the video sequence. To address this issue, \cite{tang2023temporal} propose simultaneously maintaining global referring tokens and a series of object queries. The former captures video-level referring objects based on textual expressions, while the latter is used for better localization and segmentation of objects within each frame. \cite{hu2024temporal} introduce a novel method to enhance temporal context in the segmentation of video objects, effectively improving the potential information gain of videos compared to individual images. \cite{yan2024referred} utilize collaborative cross-attention and inter-frame object correlation of visual, textual, and speech modalities to construct a unified temporal transformer for efficient referring video object segmentation. By extracting a topic-centered short text expression from the original long text expression, \cite{yuan2024losh} use both long and short text expressions for joint prediction and insert long-short cross-attention modules to joint features.
    
    Despite the relative effectiveness of current RVOS techniques, they are primarily restricted to regular supervised learning settings, which would not be able to deal with unseen scenes with few shots, not to mention multi-object cases.

\section{Methods}\label{sec:methods}
    \begin{figure*}[ht]
    \centering
    \includegraphics[width=0.99\linewidth]{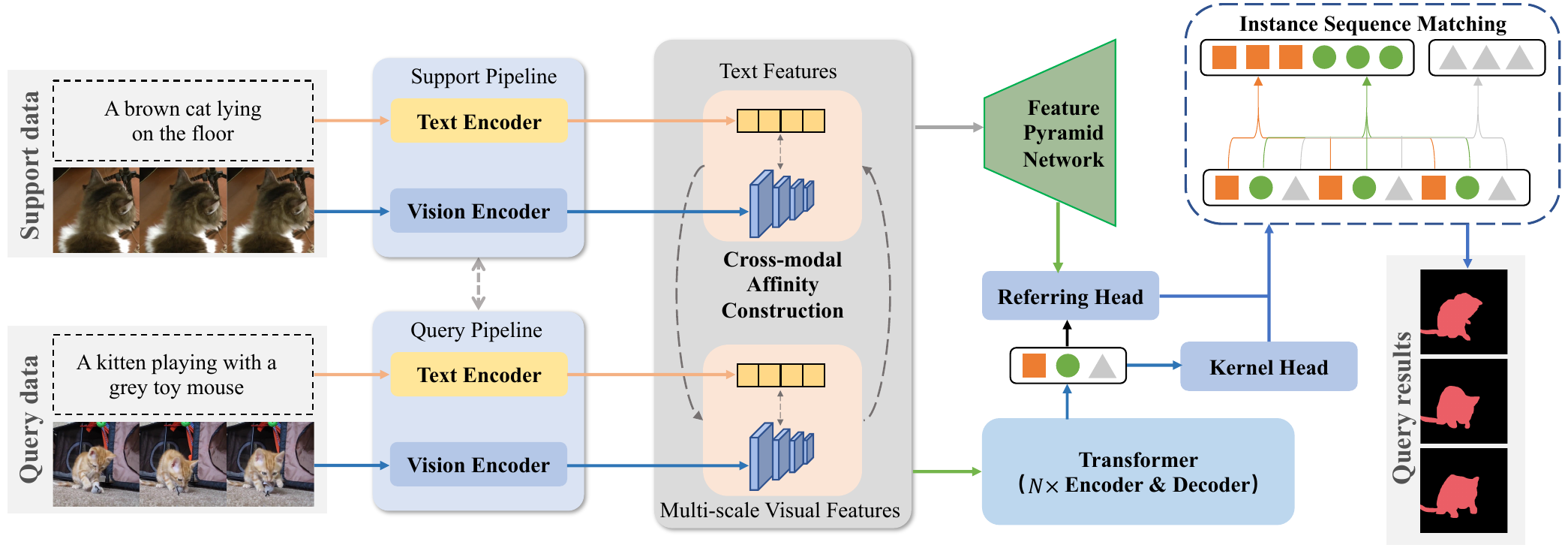}
    \vspace{-1mm}
    \caption{The overall pipeline of our framework. The feature encoder extracts visual and textual information from the support and query sets. The cross-modal affinity module calculates the multi-modal information affinity between the support set and the query set. Then, the fused features are enhanced by the transformer and are used to obtain a serial of segmentation masks through the kernel head. At the same time, the fused features are further refined across different scales by the feature pyramid network and are utilized to obtain the referring matching scores through the referring head. Finally, the multi-object segmentation result is obtained through the instance sequence matching process.}
    \label{fig:figure2}
    \vspace{-2mm}
    \end{figure*}
    \subsection{Overview} \label{ssec:overview}
    In the FS-RVOS or FS-RVMOS setting, we have disjoint training and testing datasets denoted as $D_{train}$ and $D_{test}$, each with their respective category sets $C_{train}$ and $C_{test}$, where $C_{train} \cap C_{test}$ = $\emptyset$. Similar to the few-shot learning task (\cite{snell2017prototypical}), this work adopts an episode training strategy, where $D_{train}$ and $D_{test}$ are organized into several episodes. Each episode comprises a support set $S$ and a query set $Q$, with the text-referred objects (target objects) from both sets belonging to the same class. The support set $S$ consists of $K$ image-mask pairs $S=\left\{x_{k},m_{k} \right\}^{K}_{k=1}$, along with the corresponding referring expression containing $L$ words $T_s=\left\{ t_i \right\}^{L}_{i=1}$, where $m_k$ represents the ground-truth mask of the video frame $x_k$.
    Meanwhile, the query set $Q$ comprises a selection of consecutive frames from a video, denoted as $Q=\left\{ x_{i}^{q} \right\}_{i=1}^{T}$ ($T$ represents the number of frames), along with their corresponding natural language description containing $M$ words $ T_q=\left\{ t_i \right\}^{M}_{i=1}$. With this setting, FS-RVOS or FS-RVMOS encourages the models to segment objects belonging to unseen classes in the query set over a limited number of samples provided in the support set. \par

    As shown in Figure \ref{fig:figure2}, our FS-RVMOS framework comprises several key components: a feature extraction module, a cross-modal affinity module (CMA), a Transformer, a feature pyramid network (FPN), and an instance sequence matching (ISM) process. With the support and query data as input, the pipeline predicts object masks for the query data, under the guidance of the corresponding language expressions. Specifically, the vision and text encoders extract features for visual and textual inputs, respectively. The CMA module hierarchically fuses visual and textual features, which further establishes the relationships between the support set and the query set. The fused features are then enhanced through the Transformer to capture long-range dependencies and contextual information. After passing the kernel head, a series of segmentation masks can be obtained. At the same time, the fused features are further refined 
    by the FPN to leverage cross-modal information across different spatial scales. Then, based on the refined features, the referring matching scores are obtained through the referring head. Finally, the final segmentation results can be obtained through the ISM process.
    
    Through this comprehensive pipeline, our framework effectively integrates visual and textual information to achieve accurate multi-object segmentation guided by language expressions.\par
    \subsection{Feature Extraction}\label{ssec:feature_extraction}
   We employ a shared-weight visual encoder to extract multi-scale features from each frame in both the support set and the query set, yielding visual feature sequences $F_{vs}=\left\{ f_{vs} \right\}_{vs=1} ^K$ and $F_{vq}=\left\{ f_{vq} \right\}_{vq=1}^N$. These sequences represent the visual features extracted from the support set and the query set, respectively.
    For linguistic information, we utilize a Transformer-based text encoder (\cite{liu2019roberta}) to extract text features $F_{ts}=\left\{ f_{ts} \right\}_{ts=1}^{L}$ and $F_{tq}=\left\{ f_{tq} \right\}_{tq=1}^{M}$ from the natural language descriptions $T_s$ and $T_q$. These features correspond to the linguistic inputs associated with the input support data and the query data, respectively.
    \subsection{Cross-modal Affinity Construction}\label{ssec:cma}
    In FS-RVOS, the goal is to efficiently utilize the available information from the support set and swiftly adapt to relevant scenarios, given only a few samples. Unlike conventional few-shot VOS tasks, FS-RVOS not only establishes affinity between the support and query sets but also involves multi-modal relationships between videos and referring expressions. Consequently, FS-RVOS presents unique challenges, necessitating specialized solutions for achieving high-quality results. To address these challenges, we propose the cross-modal affinity (CMA) module, as illustrated in Figure 3, where $m$ indicates the labeled ground-truth object mask in the support set.\par
\begin{figure}[t]
  \centering
  \includegraphics[width=\linewidth]{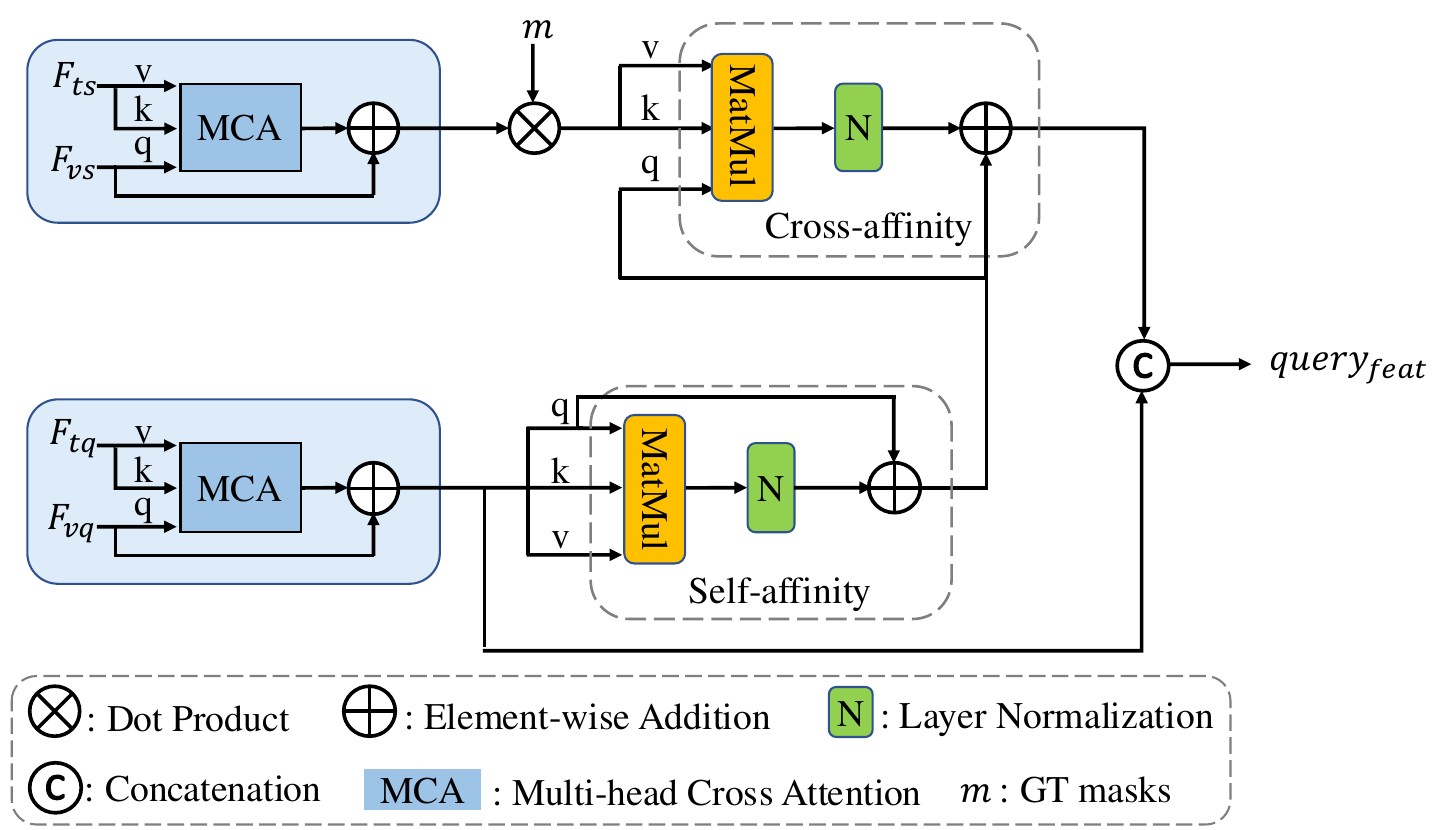}
  \vspace{-1ex}
   \caption{The architecture of the Cross-modal Affinity (CMA) module. We use multi-head cross-attention to fuse visual and text features to obtain more robust features. Self-affinity for modeling contextual information on query features and cross-affinity for aggregating beneficial information from support features.}
   \label{fig:figure3}
   \vspace{-2ex}
\end{figure}
    Specifically, the CMA module facilitates effective cross-modal and cross-data integration by exploiting the correlations between the support set and the query set, as well as the complementary strengths of visual and textual information. To this end, we employ a cross-attention mechanism to align and fuse the visual and textual features of the support and query data, respectively, generating enhanced cross-modal representations. Furthermore, we establish both self-affinity and cross-affinity relationships to enrich the query features with relevant information from the support features. These affinity relationships, modeled within the query feature context, along with the support and query features, enable efficient information exchange to optimize the query feature representations.\par
    Due to the diverse nature of referred objects and the significant variations between video frames, accurately locating the target solely through visual information poses a challenge. To enhance the precision of target segmentation, we utilize language information as a complementary source, offering specific descriptions of the referred objects.
    In order to effectively integrate and align visual and textual features, we deploy multi-head cross-attention (MCA) to fuse multi-modal information, yielding two multi-scale feature maps, $F_{vs}^{'}=\{ f_{vs}^{'} \}_{s=1}^K$ and $F_{vq}^{'}=\{ f_{vq}^{'} \}_{q=1}^N$, which defined as :
    \begin{equation}
        f_{vs}^{'} = MCA(f_{vs},f_{ts}), f_{vq}^{'} = MCA(f_{vq},f_{tq}),
    \end{equation}
    where $f_{vs}$ and $f_{vq}$ represent the visual features of the support and query data, respectively. $f_{ts}$ and $f_{tq}$ denote their corresponding textual features. By computing an affinity between textual and visual features, irrelevant visual information is filtered out. MCA is preferred over concatenation in our framework because it leverages the similarities between multi-modal features for effective information complementing. \par
    The affinity between the support set and query set serves as an indicator of the multi-modal feature correlation between them, offering valuable insights for segmenting the query data. Despite belonging to the same category, the objects segmented in the support and query sets often exhibit significant visual differences, including appearance, pose, and scene. Consequently, only a fraction of the information in the support data is conducive to effectively segmenting the query data, while the remaining information may lead to sub-optimal results. Therefore, it is imperative to accurately compute the affinity relationship between multi-modal information in the support and query sets.\par
    To tackle this issue, we propose utilizing a self-affinity block to capture and refine the query features, alongside a cross-affinity block that directs the query features to concentrate on valuable information from the support features. Specifically, given the input query features, we employ a convolution operation to map them as query $q_q$, key $k_q$, and value $v_q$. Similarly, we perform the same operation for the support features to map them to key $k_s$ and value $v_s$. The self-affinity block does not include the support features and primarily aggregates the query features, aiming to enhance segmentation accuracy. \par
    First, we compute the affinity map  $A^Q = \frac{q_q \cdot (k_{q})^{\text{T}}}{\sqrt{d_{head}}}$, where $d_{head}$ denotes the hidden dimension of the input sequences. We assume that all sequences have the same dimension, which defaults to 256. Therefore, the query features after the self-affinity block are represented as:
    \begin{equation}
        q_s = \text{Softmax}(A^Q)v_{q}.
    \end{equation} 
    Note that during CMA operation, all the features are flattened into $1D$ tensors. For example, the shape of the query features is transformed into $[B, N\times H\times W, C]$, where $B$ is the batch size, $N$ is the number of queries, $H\times W$ is the image size of one frame, $C$ is the channel size.
    \par
    We then pass the obtained query features to the cross-affinity block. The cross-affinity block aims to establish the cross-affinity relationship between the support features and the query features, aggregating useful information. Our cross-affinity block can be formulated as:
      \begin{equation}
        query_{feat} = \text{Softmax}(\frac{q_s \cdot (k_{s})^{\text{T}}}{\sqrt{d_{head}}})v_{s},
    \end{equation}\par
    \noindent where $q_s$ is the output of the self-affinity block.
    Through these two modules, query features are enhanced by modeling contextual information and computing the correlation between support and query features. \par
    Actually, this two-level structure of self-affinity and cross-affinity closely resembles the scheme of meta-learning. Specifically, the self-affinity block acts as a base learner by capturing intra-query relationships, whereas the cross-affinity block serves as a higher-level meta-learner, leveraging support-query interactions to refine the query features.

    \subsection{Mask Generation}\label{ssec:mask_generation}
    The mask generation module aims to identify relevant targets and progressively decode features. To achieve this, we leverage the structures of Deformable-DETR (\cite{zhu2020deformable}) and Feature Pyramid Networks (\cite{lin2017feature}.)
    \subsubsection{Transformer}
    Positional encodings are added to the feature sequence output by the CMA module and fed into the Transformer encoder. To enable independent processing of video frames, spatial dimensions are flattened, and the temporal dimension is shifted to the batch dimension for efficiency. The Transformer encoder's output is then passed to the decoder, where $N$ learnable anchor boxes serve as queries representing instances for each frame. These query weights are shared across video frames, ensuring flexibility for variable-length videos and robustness in tracking object instances.
    
    Text features \( F_{ts} \) and \( F_{tq} \) are duplicated $N$ times to match the number of queries and, together with the object queries, are input into the decoder. Queries leverage language expressions to locate the referred objects, ultimately generating instance embeddings as a prediction set of size $ N_q = T \times N $ ($T$ is the number of video frames). Queries retain a consistent order across frames, forming an instance sequence. Temporal consistency in segmented objects is achieved by linking corresponding queries across frames.
    \subsubsection{Feature Pyramid Network}
    In the 4-level feature pyramid network, multi-modal features from the Transformer encoder and visual encoder are combined to form hierarchical representations. Multi-head cross-modal attention enables selective reconstruction among different modalities at each level. To reduce computational complexity, multi-scale visual features are down-sampled spatially while retaining their temporal dimensions. Visual and language features are then aligned with each other via cross-attention to enhance object features for mask prediction. Finally, the fused features are up-sampled to their original resolution and sent to the auxiliary heads after Transformer decoding. The implementation is detailed in the following formula:
    \begin{equation}
        Cross(f_v^l, f_{tq}) = \text{Softmax}(\frac{f_v^l \cdot (f_{tq})^{\text{T}}}{\sqrt{d_{head}}})f_{tq},
    \end{equation}
    where $f_{tq}$ represents the textual features of the query set, and $f_v^l$, denotes the visual features at each layer. In the attention mechanism, visual features act as queries to enhance pixels strongly correlated with language expressions. Following a top-down structure of the feature pyramid network, cross-modal feature maps are up-sampled and summed. Finally, the last-layer feature map is passed through a $3\times3$ convolution layer to generate the final feature map $F_{seg} = \left\{f_{seg}^t \right\}_{t=1} ^T$, where $f_{seg}^t \in \mathbb{R}^{\frac{H}{4}\times \frac{W}{4} \times C }$.
    \subsubsection{Dynamic Convolution Filtering}
    After the Transformer decoding, auxiliary heads are introduced, including the referring head and kernel head. The referring head is a linear projection that generates confidence scores. The kernel head consists of three consecutive linear layers, generating the parameters $W=\{w_t\}_{t=1}^{N_q}$ for $N_q$ dynamic convolution kernels. These parameters form three $1\times1$ convolution layers with a channel size of $8$, serving as convolution filters. To enhance robustness, the feature map $F_{seg}$ is concatenated with the relative coordinates of each kernel. Binary segmentation masks are then generated by applying dynamic convolution filtering.
    \subsection{Instance Sequence Matching}\label{ssec:ISM} 
     Following the above processing, we employ dynamic convolution to generate object masks. The final feature maps $f_{seg}^t$ are obtained through the feature pyramid network, and the mask prediction is computed as $\hat{m}^t = \left\{ w_{t} \ast f_{seg}^t \right\}$, where $w_{t}$ represents the dynamic convolution kernel.
     
    In the Transformer decoder, $N$ learnable anchor boxes in $T$ frames are used as queries and pass through the referring head to generate a set of $N$ prediction sequences, containing $N_q = T \times N$ predicted masks. 
    Such predicted sequences can be denoted as $\hat{y} = \left\{\hat{y}_{i} \right\}_{i=1}^{N}$, where the prediction for the $i^{\mathrm{th}}$ instance is expressed as:
    \begin{equation}
    \label{eq5}
     \hat{y}_{i}=  \left\{\hat{s}_i^t, \hat{m}_i^t \right\}_{t=1}^T,
    \end{equation}
    where $\hat{s}_i^t \in \mathbb{R}^1$ represents a confidence fractional score output by the referring header, indicating the likelihood that the instance corresponds to the referred object. $\hat{m}_i^t \in \mathbb{R}^{\frac{H}{4}\times \frac{W}{4}}$ denotes the predicted segmentation mask for the instance.
    
\begin{figure}[tbhp]
        \begin{center}
            \includegraphics[width=0.5\textwidth]{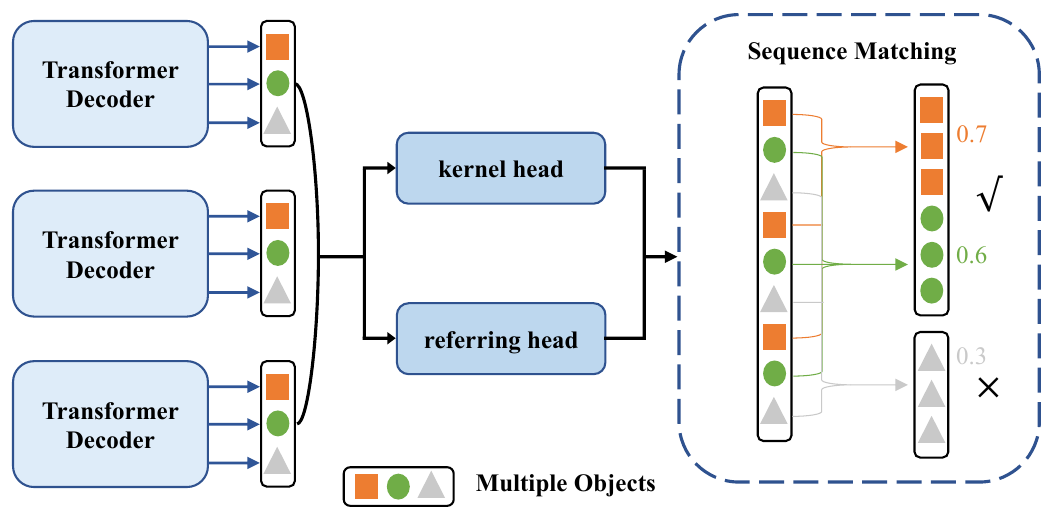}
        \end{center}
    \vspace{1ex}
    \caption{The schematic diagram of ISM for referring video multi-object segmentation.}
    \label{fig:figure35}
    \vspace{-2ex}
    \end{figure}

    Given an input video and the associated linguistic expression, our model produces $N_q$ candidate masks of $N$ instances. To refine these candidates into the final prediction, we introduce the Instance Sequence Matching (ISM) module. The ISM module evaluates the confidence scores generated by the referring head against a predefined threshold. Masks with confidence scores exceeding the threshold are retained, while those falling below are discarded. The general schematic of the ISM module for referring multi-object video segmentation is depicted in Figure \ref{fig:figure35}.

     The ISM module is highly flexible and can be seamlessly applied to referring single-object segmentation and multi-object segmentation tasks. In the context of FS-RVOS, for each frame, the instance sequence with the highest confidence score is selected as the final predicted one. Thus, the corresponding index of the chosen instance can be expressed as:
     \begin{equation}
      \label{eq6}
        {idx} = \mathop{\arg\max}_{i \in \left \{ 1, 2, ...,N \right \}}\hat{s}_{i},
    \end{equation}
    where $\hat{s}_i = \frac{1}{T}\sum_{t=1}^{T} \hat{s}_i^t$.
    
    For FS-RVMOS, where the objective is to segment multiple objects in each frame, the instance sequences with confidence scores exceeding a predefined threshold $\sigma$ are selected as predictions. The indices of the selected instances can be obtained by:
        \begin{equation}
        \label{eq7}
        {idx} = \left\{\hat{s}_i | \hat{s}_i > \sigma \right\}
    \end{equation}
    
    Thus, the final multi-object prediction, denoted as $m=\left\{\hat{m}_{idx}^t \right\}_{t=1}^T$ for each frame, is derived from the mask candidate set $\hat{m}^t$ using the instance index set ${idx}$. For example, as illustrated in Figure \ref{fig:figure35}, the video frames contain three example objects: block, circle, and triangle. Let the confidence score threshold $\sigma$ be 0.5, the predicted confidence scores for the block and circle in each frame exceed the threshold, while the score for the triangle object falls below it. Consequently, after applying the ISM module, the segmentation results include only the predicted masks of the block and circle objects. It is noteworthy that, for single-object model training, Eq. \eqref{eq5} and Eq. \eqref{eq6} are used for the matching, while Eq. \eqref{eq5} and Eq. \eqref{eq7} are deployed for matching in multi-object training. Moreover, during multi-object training, the training dataset is updated with multi-object annotation data.
    \subsection{Loss Function} \label{sec:loss}
    Let the sequence of ground-truth instances be denoted as $y= \left\{s^t, m^t \right\}_{t=1}^T$, where $s^t$ is a one-hot value with a value of $1$ if the ground-truth instance is visible in the frame and $0$ otherwise. Based on this representation, our loss function is defined as:
    \begin{equation}
    \begin{aligned}
        \mathcal{L}(y,\hat{y}_{i})  &=     
        \lambda_{cls}  \mathcal{L}_{cls}(y,\hat{y}_{i}) 
        + \lambda_{kernel} \mathcal{L}_{kernel}(y,\hat{y}_{i}),
    \end{aligned}
    \end{equation}
    where $\lambda_{cls}$, $\lambda_{kernel}$ are hyperparameters that balance the contributions of different loss components. The classification loss, $\mathcal{L}_{cls}$, is derived from focal loss (\cite{lin2017retinanet}) and is employed to supervise the prediction of the instance sequence referring results. The kernel loss, $\mathcal{L}_{kernel}$ combines DICE loss (\cite{milletari2016dice}) and binary mask focal loss to enhance the segmentation performance.\par

\section{Benchmark}\label{sec:benchmark}
    \begin{figure*}[t]
        \begin{center}
            \includegraphics[width=0.99\textwidth]{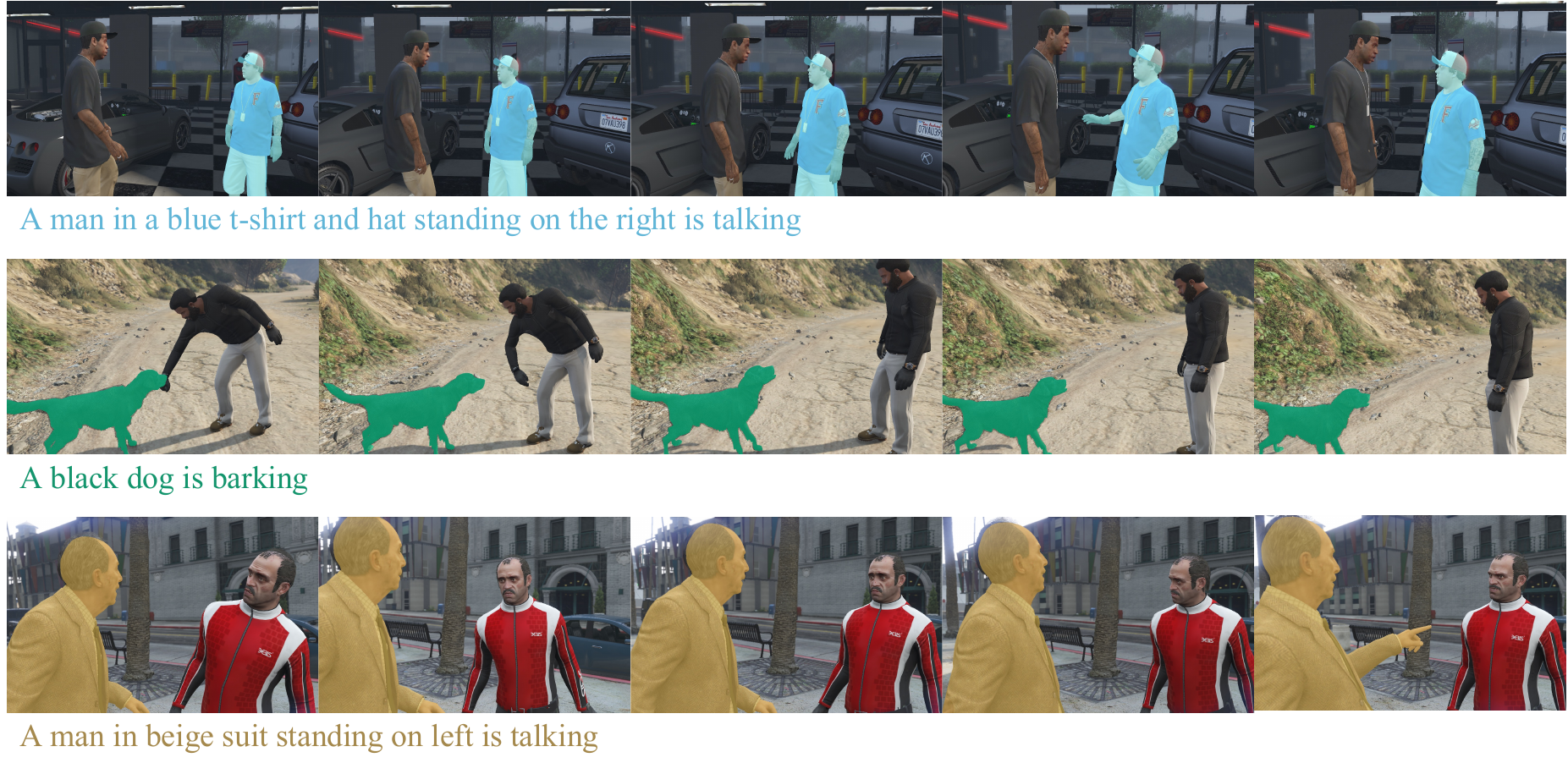}
        \end{center}
    \caption{Annotation examples of the Mini-Ref-SAIL-VOS dataset.}
    \label{fig:figure4}
    \end{figure*}
    
    Existing RVOS datasets (\cite{seo2020urvos,khoreva2018rvos,xu2015a2d,gavrilyuk2018actor}) are designed to address specific scenarios, limiting the model's ability to handle the diverse range of scenarios encountered in the real world. Furthermore, these datasets are not well suited for addressing the few-shot RVOS problem. The train/test/validation subsets within these datasets often exhibit class repetition, rendering them unsuitable for evaluating the model's performance on unseen classes and assessing its generalization capabilities.
    \subsection{Mini-Ref-YouTube-VOS}\label{ssec:mini-ytvos}
    To match the FS-RVOS setting, we built a new dataset called Mini-Ref-YouTube-VOS based on the Ref-YouTube-VOS dataset (\cite{seo2020urvos}), which contains 3,471 videos, 12,913 referring expressions, and annotated instances covering more than 60 categories.
    However, some videos in this dataset consist of multiple category instances. When preparing for the single-object few-shot setting, we cleaned up the dataset, i.e., removing such multi-category videos and keeping only those containing only one category instance, a total of 2387 videos were obtained.\par
    When constructing the Mini-Ref-YouTube-VOS dataset, we adhere to the following criteria: (1) it should encompass a broad spectrum of classes to facilitate learning generalized relationships for novel classes; (2) it should maintain class balance, ensuring each class has a similar number of samples to prevent overfitting to any specific class; (3) the classes in the training and testing sets should be mutually exclusive, facilitating the evaluation of the model's generalization to unknown classes.\par
    To maintain class balance within the dataset, we employed an intelligent sample selection strategy to address classes with limited video samples. By adjusting the number of samples, we ensured that each class had an adequate number of video samples, mitigating the risk of overfitting certain individual categories. Following this meticulous selection process, our Mini-Ref-YouTube-VOS consists of 1668 videos covering 48 object categories, with each video containing approximately 90 to 180 frames. Spanning from various organisms to daily necessities, the Mini-Ref-YouTube-VOS dataset authentically reflects the rich diversity of the real world. To better show the model results, we adopt the cross-validation method to divide the dataset into four folds on average. Each fold contains 36 training and 12 test classes with disjoint categories.
    
    \subsection{Mini-Ref-SAIL-VOS}\label{ssec:mini-sailvos}
    Most Mini-Ref-YouTube-VOS data involve natural scenes of a relatively homogeneous type and, therefore, do not represent the diversity of data in the real world. To better showcase the generalization capability of our model, we collected videos from the SAIL-VOS dataset (\cite{hu2019sail}) to create a new dataset, Mini-Ref-SAIL-VOS. The uniqueness of the SAIL-VOS dataset lies not only in its origin from the GTA-V game but also in the distinct differences between this synthetic dataset and natural scene datasets. In gaming environments, dynamic city streets, fictional buildings, and artificially designed obstacles are commonplace, presenting a stark contrast to the natural scenes encountered in the real world. This synthetic nature of the data provides a challenging, diverse, and unconventional testing environment. \par

    In SAIL-VOS, each frame is accompanied by dense annotations with semantic labels, pixel-level, and non-modal segmentation masks. Phenomena such as camera transitions, object segmentation across frames, and object occlusion are inevitable in the SAIL-VOS dataset, which brings challenges to the segmentation task.\par

    During the reorganization of the SAIL-VOS dataset, we considered multiple factors to ensure its suitability for the FS-RVOS setting. Firstly, videos with fewer segmented target frames were discarded as they might not provide sufficient data support and would be less effective for few-shot learning tasks. Secondly, to maintain the temporal continuity of segmented targets, frames without any target appearances in between were manually removed. This operation helps maintain the motion trajectories of targets in the dataset, enabling the model to understand the spatio-temporal variations of the targets better. Phenomena such as object occlusion were treated as a challenge rather than deleting related video frames. This decision was made because real-world applications often encounter partially occluded targets, and retaining these frames in the dataset makes it more realistic, simulating situations present in real-world scenes.\par

    The Mini-Ref-SAIL-VOS dataset is relatively small, comprising 68 videos and 3 semantic categories. Despite its small scale, we carefully selected these videos and categories to ensure the representativeness and challenge of the data for the FS-RVOS task. This dataset covers common semantic categories in synthetic scenes, providing an intriguing dataset for our research to validate the generalization capability of our model further. \par
   \begin{figure*}[th]
        \begin{center}
            \includegraphics[width=0.99\textwidth]{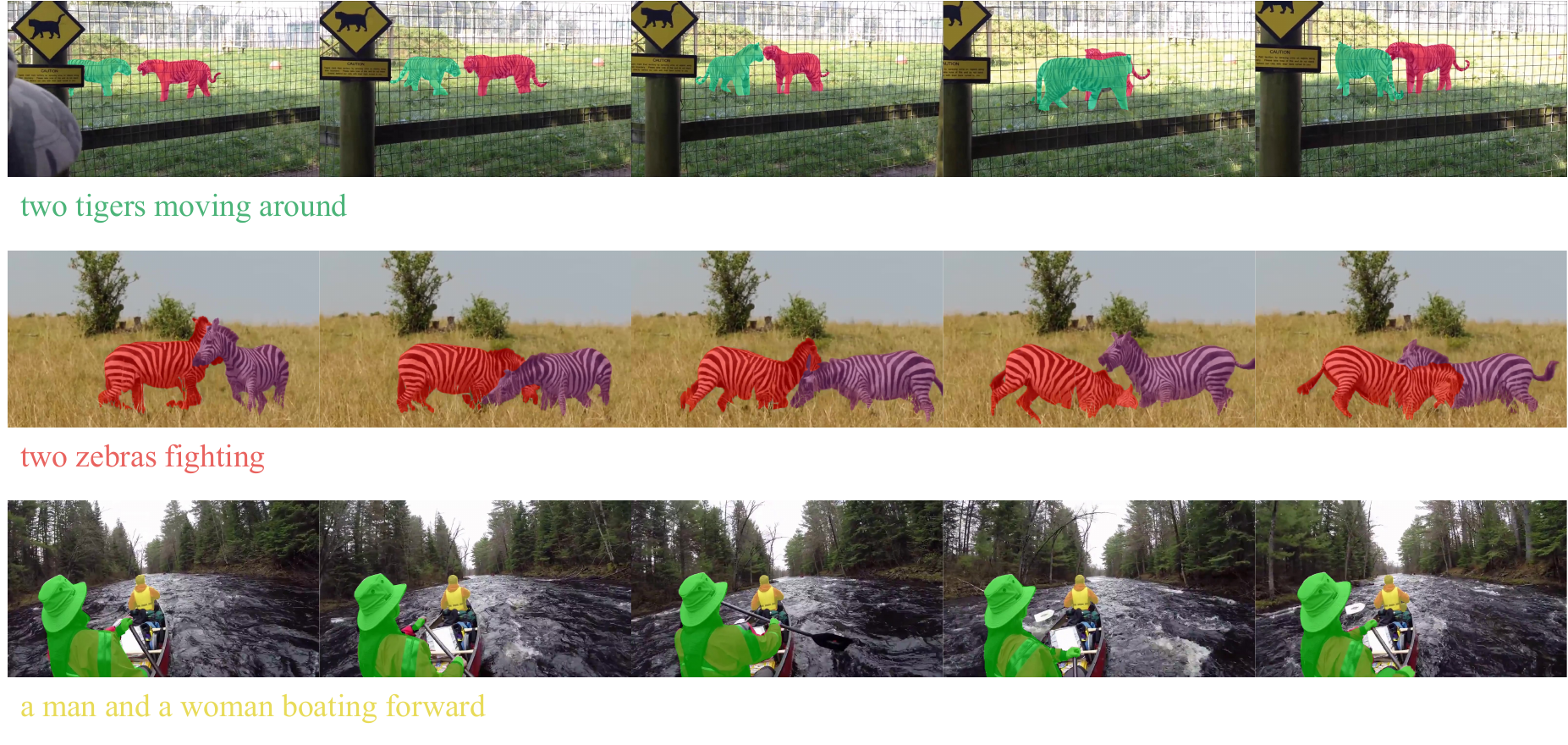}
        \end{center}
    \caption{Annotation examples of the Mini-MeViS dataset.}
    \label{fig:figure5}
    \end{figure*}
    
    It is worth noting that although there are accurate mask annotations in the SAIL-VOS dataset, natural language descriptions corresponding to the segmentation object are not available. Thus, to adapt it to FS-RVOS, we employed expert annotators to provide referring expressions after data collection. Given a pair of videos for each annotator, the video frames are superimposed with corresponding masks to indicate the objects to be segmented.
    The annotators were then asked to provide a distinguishing statement with a word limit of 20 words. To ensure the quality of natural language annotations, all annotations are verified and cleaned up after the initial annotation. The target will not be used if the natural language description cannot clearly describe the target. \par

     As shown in Figure\ref{fig:figure4}, we show some selected videos along with referring expressions. In general, the Mini-Ref-SAIL-VOS dataset evaluates model performance and comprehensively considers model generalization, adaptability, and practicality. 

    \subsection{Mini-MeViS}\label{ssec:mini-mevis}
    While previous datasets for referring video object segmentation mainly focused on single-object scenarios, failing to capture the complexities of multi-object tasks, \cite{ding2023mevis} introduced the MeViS dataset. This extensive dataset comprises 2006 videos and 8171 objects, along with 28570 motion expressions indicating these objects. MeViS emphasizes temporal motion information in videos, allowing language expressions to describe motion objects spanning random frames, capturing both transient and prolonged movements occurring throughout the entire video. Unlike existing datasets that primarily concentrate on single-object expressions, MeViS extends this task to multi-object expressions, making it more challenging and reflective of real-world scenarios.\par
    To suit the setting of few-shot referring video multi-object segmentation tasks, we created a new dataset named Mini-MeViS based on MeViS. While MeViS contains numerous videos, some include instances of multiple categories, making them unsuitable for few-shot learning settings. Additionally, incomplete annotation information in MeViS necessitated data cleansing and augmentation of category information. \par
    During the construction of the Mini-MeViS dataset, we carefully considered multiple factors to ensure its suitability for few-shot multi-object referring video segmentation. Firstly, we filtered out videos with textual descriptions corresponding to multi-objects, retaining only those with single objects. Since MeViS does not annotate class information for segmented objects, determining object categories based on textual descriptions was necessary. Considering the constraints of few-shot learning on categories, we remove the videos containing instances of multiple categories, ultimately resulting in a dataset comprising 79 videos and 5 semantic categories for Mini-Mevis dataset. This ensured that the Mini-MeViS dataset was well-suited for few-shot learning tasks. Figure \ref{fig:figure5} illustrates the selected videos and partial examples of textual expressions.\par

   \begin{figure*}[tbhp]
        \centering
        \subfloat[Two standing workers]{
		\includegraphics[width=4.8in,height=1.8in]{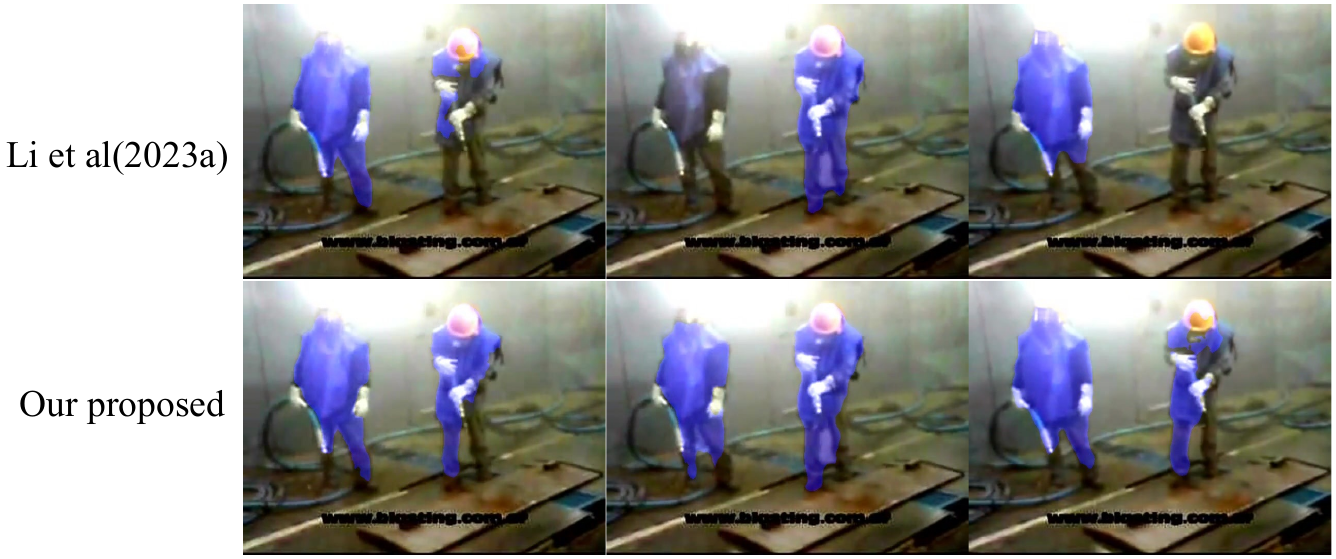} 
	}
    \\
       \subfloat[The left two giraffes standing]{
		\includegraphics[width=4.8in,height=1.8in]{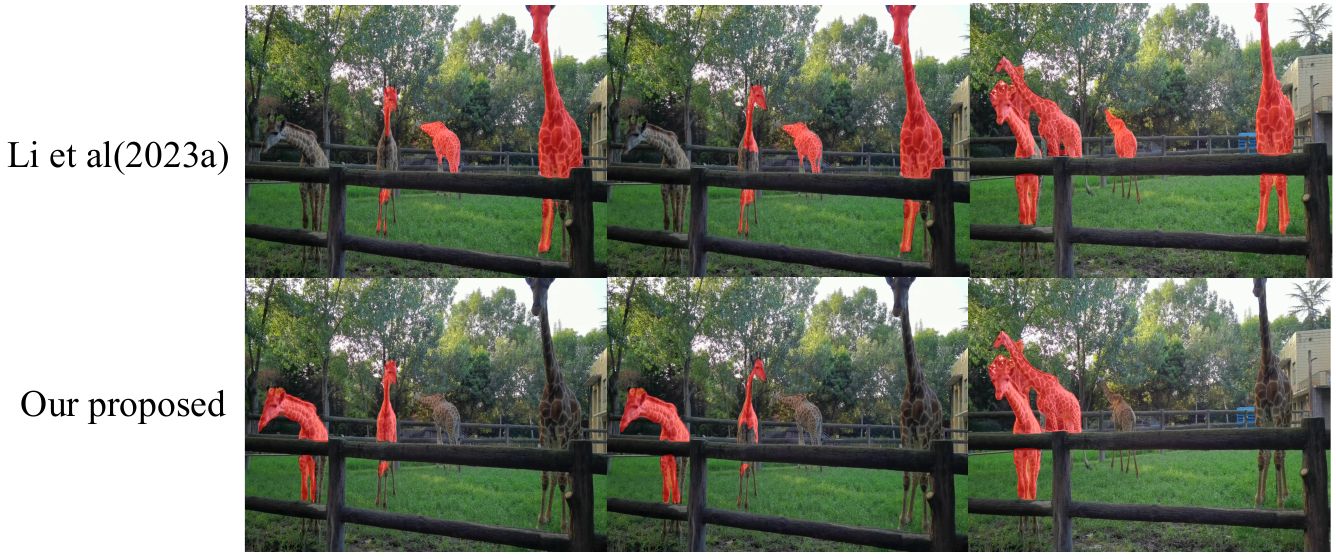}
	}
     \vspace{-1ex}
    \caption{Comparisons of the segmentation performance of our proposed approach and the method from  \cite{li2023learning} across various video scenes. The comparisons highlight the robustness of our approach under diverse conditions.}
    \label{fig:figure6}
    \vspace{-3ex}
    \end{figure*}
\section{Experiments}\label{sec:exp}
    \subsection{Implementation details}\label{ssec:implementation}
    We adopt ResNet-50 (\cite{he2016resnet}) and RoBERTa-Base (\cite{liu2019roberta}) as our vision and text encoders, respectively. During the training stage, the parameters of both encoders are frozen. In our experiments, we adopted a 5-shot setting. Specifically, we extracted five consecutive frames and their corresponding referring expressions from a certain video of a particular class as the support set. The query set consisted of consecutive frames and corresponding referring expressions extracted from other videos belonging to the same class. To ensure the reliability of the results, we conducted five tests for each fold and reported the average results. Our model was optimized using Adam with a weight decay of $5 \times 10^{-4}$ and an initial learning rate of $1\times 10^{-4}$. 
    
    To balance GPU memory efficiency, we down-sampled all video frames and restricted the size of video frames, with a minimum size of 360 and a maximum size of 640. This strategy ensured that the model training process remained efficient even with limited hardware resources, enhancing the scalability of the experiments. The parameters of the loss function were set as $\lambda_{cls}=2$, $\lambda_{kernel}=5$. In the matching module, we put the matching threshold $\sigma=0.5$. All methods were pre-processed and fine-tuned in the manner of previous similar tasks, i.e., pretrained on the Ref-COCO (\cite{yu2016refcoco}) dataset. For our pipeline, we utilize the temporal information implicitly. Specifically, during training and inference phases, instead of taking individual frames, we treat all input video frames as a whole and predict the mask trajectory for the entire video with only one forward pass.
    Following the settings of previous RVOS works, we use the region similarity ($\mathcal{J}$) and the contour accuracy ($\mathcal{F}$) to measure the model performance. 
    \subsection{Results}\label{ssec:results}
    As a novel task, FS-RVOS lacks directly comparable related works. Therefore, considering our mutual focus on few-shot video segmentation, we adopt DANet (\cite{chen2021delving}) as the baseline. For a fair comparison, we add a visual-language fusion module to the Few-Shot VOS model.

    Firstly, to demonstrate the performance difference from our previous work (\cite{li2023learning}), we compare it with our proposed approach in this work by conducting diverse segmentation tests on different scene videos, as shown in Figure\ref{fig:figure6}. The figure shows that when dealing with these multi-object videos, our previous method (\cite{li2023learning}) is unstable, either unable to cover the referring multiple objects, such as `workers' in the first row of Figure \ref{fig:figure6}(a), or segment many semantically irrelevant targets, such as `giraffes' in the first row of Figure \ref{fig:figure6}(b). Whereas, due to the organic integration of many merit mechanisms during mask generation, such as multi-scale visual-linguistic feature fusion, dynamic convolution perception and multi-instance sequence matching, our proposed approach can more wholly and accurately segment the target objects according to the reference text, demonstrating its superiority and better scene adaptability compared to our previous work (\cite{li2023learning}). Then, to demonstrate the performance of our approach in more real-world scenarios, we also conduct rich referring segmentation experiments on three constructed datasets (Mini-Ref-YouTube-VOS, Mini-Ref-SAIL-VOS, and Mini-MeViS) and make detailed quantitative analysis, which will be introduced one by one below.
\begin{table*}[tbhp]
    \begin{center}
        \caption{Quantitative results on Mini-Ref-YouTube-VOS.}
        \label{tab:table1}
        \begin{tabular}{l|c|c|c|c|c|c}
    \hline
    \multicolumn{2}{c|}{\multirow{1}{*}{Method}} & \multicolumn{1}{c|}{Fold-1} & \multicolumn{1}{c|}{Fold-2} & \multicolumn{1}{c|}{Fold-3} & \multicolumn{1}{c|}{Fold-4} & \multicolumn{1}{c}{Mean}\\
    
    \hline
    
    \multirow{3}{*}{$\mathcal{J}$} 
    & DANet~\cite{chen2021delving} & {47.0} & 33.5 & 38.5 & {44.0} & 40.8  \\
    & HPAN~\cite{tang2024holistic} & {58.5} & 41.1 & \textbf{52.4} & {52.5} & 51.1  \\
    \multirow{3}{*}{} 
    & Ours & \textbf{59.5} & \textbf{45.3} & {50.8} & \textbf{57.3} & \textbf{53.1}  \\

    \hline

    \multirow{3}{*}{$\mathcal{F}$} 
    & DANet~\cite{chen2021delving}  & 49.3 & 38.2 & 41.4 & 45.8 & 43.7  \\
    & HPAN~\cite{tang2024holistic}  & 54.3 & 42.3 & 49.8 & 56.3 & 50.7  \\
    \multirow{3}{*}{} 
    & Ours  & \textbf{60.8} & \textbf{48.9} & \textbf{51.3} & \textbf{58.1} & \textbf{54.8}  \\

    \hline
    \end{tabular}
    \end{center}
    \vspace{-1ex}
\end{table*}

\begin{table*}[tbhp]
\vspace{-1ex}
    \begin{center}
        \caption{Quantitative results on Mini-Ref-SAIL-VOS}
        \label{tab:table2}
        \begin{tabular}{l|c|c|c|c|c|c}
    \hline
    \multicolumn{2}{c|}{\multirow{1}{*}{Method}} & \multicolumn{1}{c|}{Fold-1} & \multicolumn{1}{c|}{Fold-2} & \multicolumn{1}{c|}{Fold-3} & \multicolumn{1}{c|}{Fold-4} & \multicolumn{1}{c}{Mean}\\
    
    \hline
    
    \multirow{3}{*}{$\mathcal{J}$} 
    & DANet~\cite{chen2021delving}  & 54.3 & 47.6 & 30.9 & 30.6 & 40.9  \\
    & HPAN~\cite{tang2024holistic} & {68.6} & 72.0 & 70.7 & \textbf{73.6} & 71.2  \\
    \multirow{2}{*}{} 
    & Ours & \textbf{80.9} & \textbf{80.8} & \textbf{80.1} & {68.9} & \textbf{77.7}  \\
    
    \hline
    \multirow{3}{*}{$\mathcal{F}$} 
    & DANet~\cite{chen2021delving}  & 54.1 & 48.4 & 35.2 & 36.0 & 43.4  \\
    & HPAN~\cite{tang2024holistic} & {61.7} & 71.1 & 66.2 & \textbf{71.6} & 67.7  \\
    \multirow{2}{*}{} 
    & Ours  & \textbf{77.1} & \textcolor{blue}{\textbf{77.0}} & \textbf{77.3} & {67.8} &\textbf{74.8}  \\

    \hline
    \end{tabular}
    \end{center}
    \vspace{-1ex}
\end{table*}

        \subsubsection{Mini-Ref-YouTube-VOS}\label{subsubsec:mini-ref-youtubevos}
        The comparison experimental results of our model with DANet (\cite{chen2021delving}) and another cutting-edge few-shot work, HPAN (\cite{tang2024holistic}), on the Mini-Ref-YouTube-VOS dataset are demonstrated in Table \ref{tab:table1}. We compare the four folds of the Mini-Ref-YouTube-VOS dataset and average the outputs across these folds to represent the model's comprehensive performance. As shown in Table 1, our proposed method achieves performance improvement, with an average gain exceeding 10\% over DANet (\cite{chen2021delving}) and 2\% over HPAN (\cite{tang2024holistic}) in terms of region similarity (J) and contour accuracy (F). These consistent improvements across a variety of videos underscore the robustness of our approach in addressing the challenges inherent to the Mini-Ref-YouTube-VOS dataset. The results highlight our method's ability to deliver significant advances in performance, providing compelling evidence of its practical applicability in real-world scenarios. Furthermore, these findings reinforce the superiority of our approach in the domain of object segmentation.
        \subsubsection{Mini-Ref-SAIL-VOS}\label{subsubsec:mini-ref-sailvos}
        To further assess the generalization capability of our model, we have conducted detailed experiments and comparisons on the Mini-Ref-SAIL-VOS dataset. It is noteworthy that we do not train the model anew on the Mini-Ref-SAIL-VOS dataset. Instead, we directly utilize the models trained on the four different Folds of the Mini-Ref-YouTube-VOS dataset for testing. It should be noted that the videos in the Mini-Ref-SAIL-VOS dataset are from gaming scenes, significantly different in scene domain from the data in the Mini-Ref-YouTube-VOS dataset. Additionally, some objects in the videos may be occluded, adding to the dataset's challenges. The comparison results in Table \ref{tab:table2} demonstrate that our method outperforms both DANet (\cite{chen2021delving}) and HPAN (\cite{tang2024holistic}). These results highlight the robustness and efficacy of our approach, especially in processing video data from gaming scenes and handling challenging conditions, such as object occlusion and fast motion. This robust performance across varied domains and complex scenarios emphasizes our method's generalization capability, making it a potentially promising solution for practical applications such as dynamic scene understanding. \par
        
        In addition, we conduct a comparison between our method and some state-of-the-art (SOTA) RVOS approaches, including LBDT (\cite{ding2022language}), ReferFormer (\cite{wu2022language}), MTTR (\cite{botach2022end}) and MUTR (\cite{yan2024referred}), with the results presented only for Fold-1. We first test them directly on the Mini-Ref-SAIL-VOS dataset. Subsequently, for a more fair comparison, we fine-tune the models with a small number of samples to simulate the few-shot learning setting. For models that support multiple backbones, we ensure consistency by selecting ResNet50, the same backbone used in our method. The corresponding results are shown in Table \ref{tab:table3}, with the underlined values indicating the performance after fine-tuning. It is clear that, while fine-tuning improves the performance of these models, there remains a substantial performance gap between them and our approach. This disparity is due to our model's ability to effectively establish multi-modal affinity relationships between the support and query sets, allowing for rapid adaptation to new scenes with minimal samples, leading to superior performance.
\begin{table*}[!t]
\vspace{-1ex}
    \begin{center}
        \caption{Comparison with the state-of-the-art RVOS methods on the Mini-Ref-SAIL-VOS dataset to evaluate the model’s generalization. Underlined scores are achieved after fine-tuning.}
        \label{tab:table3}
        \begin{tabular}{c | c | c |c}
    \hline

    \multirow{1}{*}{Method} & \multirow{1}{*}{$\mathcal{J}$} & \multirow{1}{*}{$\mathcal{F}$} & \multirow{1}{*}{$\mathcal{J} \& \mathcal{F}$}  \\

    \hline
    \multirow{1}{*}{LBDT~\cite{ding2022language}} 
     & 27.5 / \underline{42.4} & 36.2 / \underline{37.3} &  31.6 / \underline{39.6}\\

    \hline

    \multirow{1}{*}{ReferFormer~\cite{wu2022language}} & 65.1 / \underline{74.1} & 62.8 / \underline{64.9} & 64.0 / \underline{69.5}\\

    \hline

    \multirow{1}{*}{MTTR~\cite{botach2022end}} & 66.5 / \underline{69.7} & 64.9 / \underline{68.1} & 65.7 / \underline{68.9}\\

    \hline

    \multirow{1}{*}{MUTR~\cite{yan2024referred}} & 66.0 / \underline{74.3} & 70.8 / \underline{71.5} & 68.4 / \underline{72.9}\\

    \hline

    \multirow{1}{*}{Ours} & \textbf{80.9} & \textbf{77.1} & \textcolor{blue}{\textbf{79.0}} \\

    \hline

\end{tabular}
    \end{center}
    \vspace{-1ex}
\end{table*}
\begin{table*}[!t]
\vspace{-1ex}
    \begin{center}
        \caption{Quantitative results on Mini-MeViS}
        \label{tab:table4}
        \begin{tabular}{l|c|c|c|c|c|c}
    \hline
    \multicolumn{2}{c|}{\multirow{1}{*}{Method}} & \multicolumn{1}{c|}{Fold-1} & \multicolumn{1}{c|}{Fold-2} & \multicolumn{1}{c|}{Fold-3} & \multicolumn{1}{c|}{Fold-4} & \multicolumn{1}{c}{Mean}\\
    
    \hline
    
    \multirow{3}{*}{$\mathcal{J}$} 
    & DANet~\cite{chen2021delving}  & 25.3 & 25.9 & 24.0 & 24.3 & 24.9  \\
    & HPAN~\cite{tang2024holistic} & {28.0} & 29.1 & 28.9 & 26.4 &   28.1\\
    \multirow{3}{*}{} 
    & Ours & \textbf{33.6} & \textbf{34.4} & \textbf{35.4} & \textbf{32.0} & \textbf{33.9}  \\
    
    \hline
    \multirow{3}{*}{$\mathcal{F}$} 
    & DANet~\cite{chen2021delving}  & 32.9 & 33.8 & 32.8 & 32.8 & 33.1  \\
    & HPAN~\cite{tang2024holistic} & {35.0} & 36.7 & {35.1} & 34.3 & 35.3\\
    \multirow{3}{*}{} 
    & Ours  & \textbf{39.9} & \textbf{39.8} & \textbf{41.9} & \textbf{37.9} &\textbf{39.9}  \\

    \hline
\end{tabular}
    \end{center}
    \vspace{-1ex}
\end{table*}
        \subsubsection{Mini-MeViS}\label{subsubsec:mini-mevis}
        The comparison experimental results on the Mini-MeViS dataset are presented in Table \ref{tab:table4}. As shown in Table \ref{tab:table4}, our method outperforms both DANet (\cite{chen2021delving}) and HPAN (\cite{tang2024holistic}), with notable improvements in segmentation accuracy. A key advantage of our approach is the proposed ISM module, which operates without the need for retraining and can be applied directly during testing. This eliminates the need for domain-specific fine-tuning, allowing us to use models pre-trained on the Mini-Ref-YouTube-VOS dataset for testing on the Mini-MeViS dataset. This design not only enhances the flexibility of our approach, but also demonstrates its robust generalization capability. The experimental results highlight our method’s ability to effectively handle multi-object scenarios, as well as its strong generalization on new, unseen datasets, making it a reliable solution for practical applications in dynamic and complex environments.\par

        To further evaluate the performance of our model, we compare it with several SOTA RVOS methods, including LBDT (\cite{ding2022language}), ReferFormer (\cite{wu2022language}) and MUTR (\cite{yan2024referred}), by testing all approaches directly on the Mini-MeViS dataset. Thus, to observe our model across a broader range of scenarios, we fine-tune it using additional data. Specifically, we select 18 videos from 5 categories in the Mini-MeViS dataset for fine-tuning (containing 6 epochs of training), with a particular emphasis on the 'person' category as humans are one of the most common and essential objects in multi-object referring video object segmentation tasks. By leveraging such diverse data for fine-tuning, we aim to assess our model's adaptability to various scenes in real complex environments. \par
        
        Actually, Table \ref{tab:table4} reveals that for all FS-VOS methods, testing on Fold-3 of the Mini-MeViS dataset yields the best results, while testing on Fold-4 yields the worst results. To further investigate performance, we compare our model with other SOTA RVOS methods in Fold-3 and Fold-4, with the results presented in Table \ref{tab:table5}. Since MUTR (\cite{yan2024referred}) does not release the pre-trained model, we train it on our Mini-Ref-YouTube-VOS dataset and then test it on the Mini-MeViS dataset. As shown in Table \ref{tab:table5}, the underlined values represent the results after fine-tuning (using the same fine-tuning approach as ours). According to the table, the performance of MUTR (\cite{yan2024referred}) is inferior to ReferFormer (\cite{wu2022language}). This may be because its model is too complex and insufficiently trained, making applying for few-shot multi-object video segmentation challenging. Our method outperforms all others, demonstrating the effectiveness of the ISM module, which efficiently selects optimal trajectories and improves multi-object tracking accuracy by utilizing complementary information from different trajectories. These results not only confirm the effectiveness of our approach but also highlight its potential for advancing few-shot referring video segmentation in future studies.
\begin{table*}[!t]
\vspace{-1ex}
    \begin{center}
        \caption{Comparison with the SOTA RVOS methods on the Mini-MeViS dataset to assess the model’s generalization.}
        \label{tab:table5}









\begin{tabular}{c | c | c |c}
    \hline

    \multirow{1}{*}{Method} & \multirow{1}{*}{$\mathcal{J}$} & \multirow{1}{*}{$\mathcal{F}$} & \multirow{1}{*}{$\mathcal{J} \& \mathcal{F}$}  \\

    \hline
    \multirow{1}{*}{LBDT~\cite{ding2022language}} 
     & 16.9 / \underline{21.2} & 26.5 / \underline{30.5} &  21.7 / \underline{25.9}\\

    \hline

    \multirow{1}{*}{ReferFormer~\cite{wu2022language}} & 34.8 / \underline{36.7} & 41.5 / \underline{44.5} & 38.2 / \underline{40.6}\\

    \hline

    \multirow{1}{*}{MUTR~\cite{yan2024referred}} & 29.5 / \underline{33.6} & 34.7 / \underline{40.7} & 32.1 / \underline{37.2}\\

    \hline

    \multirow{1}{*}{Ours(Fold4)} & 32.0 / \underline{36.6} & 37.9 / \underline{44.5} & 35.0 / \underline{40.6}\\

    \hline

    \multirow{1}{*}{Ours(Fold3)} 
    & \textbf{35.4} / \underline{\textbf{37.9}} & \textbf{41.9} / \underline{\textbf{45.3}} 
    & \textbf{38.7} / \underline{\textbf{41.6}}\\

    \hline

\end{tabular}

    \end{center}
    \vspace{-1ex}
\end{table*}
\begin{table}[t]
\vspace{-1ex}
\tabcolsep=0.5cm 
     \caption{Ablation studies that validate the effectiveness of each component in our CMA. The first result is obtained with our baseline.}
     \label{tab:table6}
    \begin{tabular}{c | c | c}

\hline
\multirow{1}{*}{Self-affinity} & \multirow{1}{*}{Cross-affinity} & \multirow{1}{*}{$\mathcal{J} \& \mathcal{F}$}  \\
\hline

-\ & -\ & 57.9\\
\hline

\multirow{1}{*}{\checkmark} & -\ & 59.2\\
\hline

\multirow{1}{*}{\checkmark} & \multirow{1}{*}{\checkmark} & \textbf{60.2}\\
\hline

\end{tabular}

    \vspace{-2ex}
\end{table}
\begin{table}[t]
\vspace{-2ex}
\tabcolsep=0.34cm 
     \caption{Ablation studies on Textual Expression.}
     \label{tab:table65}
    
		\begin{tabular}{c c c c c}
			\hline
			&\multicolumn{2}{c}{W/O Text} & \multicolumn{2}{c}{With Text} \\
			\hline
			Unsee Category & $\mathcal{J}$ & $\mathcal{F}$ & $\mathcal{J}$ &  $\mathcal{F}$\\
			\hline
			Bottle & 77.3 & 51.6 & \textbf{82.4} & \textbf{56.3}\\
			
			Coffee & 74.7 & 64.3 & \textbf{81.3} & \textbf{71.0} \\ 
			
			Lego & 69.5 & 66.7 & \textbf{84.2} & \textbf{80.7}\\
			
		    Whisky& 90.4 & 91.8 & \textbf{95.2} & \textbf{96.4}\\
			
			Oven & 65.0 & 31.4 & \textbf{68.4} & \textbf{35.6} \\
			
			Plush Toys & 69.2 & 65.8 & \textbf{78.0} & \textbf{68.9}\\
			
			\hline
		\end{tabular}

    \vspace{-1ex}
\end{table}
%
\begin{table}[t]
\vspace{-1ex}
\tabcolsep=0.5cm 
     \caption{Ablation studies that validate the effectiveness of the Instance Sequence Matching.}
     \label{tab:table7}
    
\begin{tabular}{c | c | c |c}
    \hline

    \multirow{1}{*}{Method} & \multirow{1}{*}{$\mathcal{J}$} & \multirow{1}{*}{$\mathcal{F}$} & \multirow{1}{*}{$\mathcal{J} \& \mathcal{F}$}  \\




    \hline

    \multirow{1}{*}{w/o(Fold3))} & 34.9 & 41.4 & 38.2 \\

    \hline

    \multirow{1}{*}{w(Fold3)} 
    & \textbf{35.4} & \textbf{41.9} &  \textbf{38.7} \\

    \hline

\end{tabular}
    \vspace{-1ex}
\end{table}
    \subsection{Ablation Study}\label{ssec:ablation_study}
    \subsubsection{Cross-modal Affinity}
    In this section, we perform an ablation study on the Mini-Ref-YouTube-VOS dataset to evaluate the design and robustness of the model. Unless otherwise stated, we show results only under Fold-1. Denoting $\mathcal{J} \& \mathcal{F}$ as the average of J and F, we can use the indicator to show the performance of the model. We perform ablation studies on the components of the CMA in Table \ref{tab:table6}. The results of the baseline are shown in the first line. 
    
    As mentioned above, the baseline refers to directly concatenating the support features and query features into the FPN to obtain the segmentation mask. First, we only utilize the self-affinity block to establish contextual information between pixels to enhance query features. At this time, support features are concatenated with the enhanced query features and then fed to the Transformer. It can be seen from Table \ref{tab:table6} that quite good results can be achieved, indicating that the Transformer works to model features and extract contextual information. By adding the proposed cross-affinity block, the performance can be further improved by 1\%. Such comparisons show that the cross-affinity block correctly constructs the multi-modal relationship between the support set and the query set, effectively avoiding the bias of the query features.
    \subsubsection{Textural Expression}
     In this section, we perform ablation studies on textual expression. To better analyze the impact of text expression on the model, we collect videos of some categories that the model has not encountered before. Specifically, we select videos of six object categories that the model has never seen before in real-world scenarios. These videos comprise complex scene conditions, such as lighting changes (like cakes) and occlusions (like Lego). Then, we add textual descriptions to these video sequences through a similar approach to create the Mini-Ref-SAIL-VOS dataset. In Table \ref{tab:table65}, ‘W/O Text’ refers to experiments without textual information, while ‘With Text’ represents the use of textual expression. The results reveal that our model can achieve a comparable segmentation prediction even with only a few visual samples (without textual expression). 
     
     However, textual expressions significantly improve the segmentation accuracy, particularly in the cases where the pixel features of the target images are affected. This demonstrates the critical role of textual expressions and further highlights the applicability of our model to various unseen and complex real-world video scenarios.
    \subsubsection{Instance Sequence Matching}
    In this section, we focus on the ablation study of ISM modules using the Mini-MeViS dataset. The quantitative experimental results on Fold-3 are shown in
    Table \ref{tab:table7}.
    In the table, "w/o" indicates the absence of the instance sequence matching module, while "w" denotes its presence. It is evident from the table that the model's performance improved with the inclusion of the matching module. Table \ref{tab:table7} indicates that the ISM module successfully leveraged multiple trajectories for complementary information, enhancing the model's effectiveness in multi-object segmentation tasks. Specifically, the ISM module aids in identifying and selecting the best-matching trajectories, thereby refining the multi-object segmentation results. This result further validates the effectiveness and importance of the ISM module in enhancing model performance.\par
\begin{figure}[b]
    \centering
    \includegraphics[width=\linewidth]{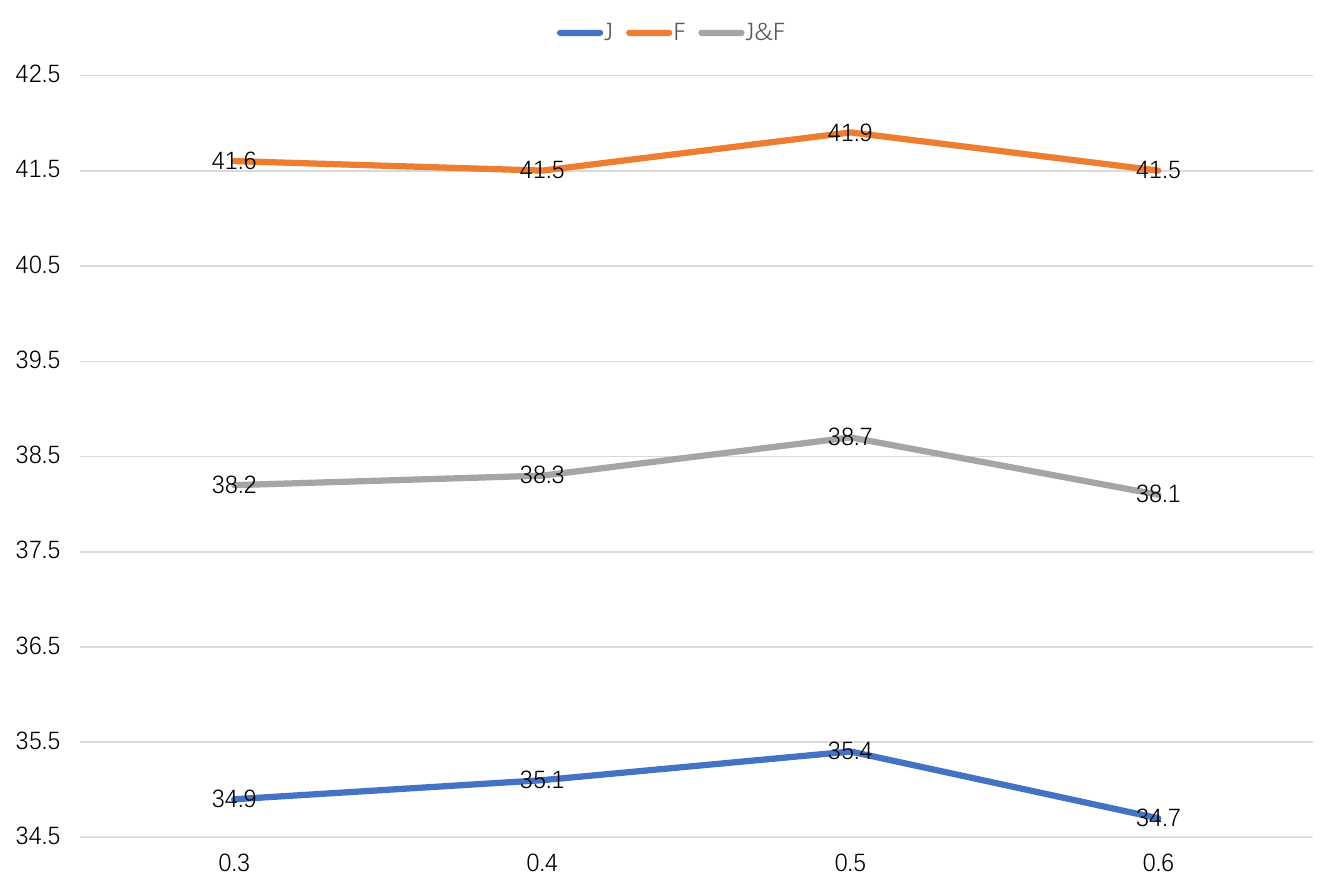}
    \vspace{1ex}
    \caption{Ablation studies that validate the influence of different matching thresholds for our model}
    \label{fig:figure7}
    \end{figure}

    Finally, we conduct a study on the impact of different matching thresholds $\sigma$ on our model's performance. As depicted in Figure \ref{fig:figure7}, the horizontal axis represents different matching thresholds, while the vertical axis represents the model's performance. It can be observed that the model's performance reaches its peak at $\sigma=0.5$. Overall, the variation in model performance is stable across different thresholds. We select $\sigma=0.5$ as the default value.

    \begin{figure*}[ht]
    \centering
    \includegraphics[width=0.9\linewidth]{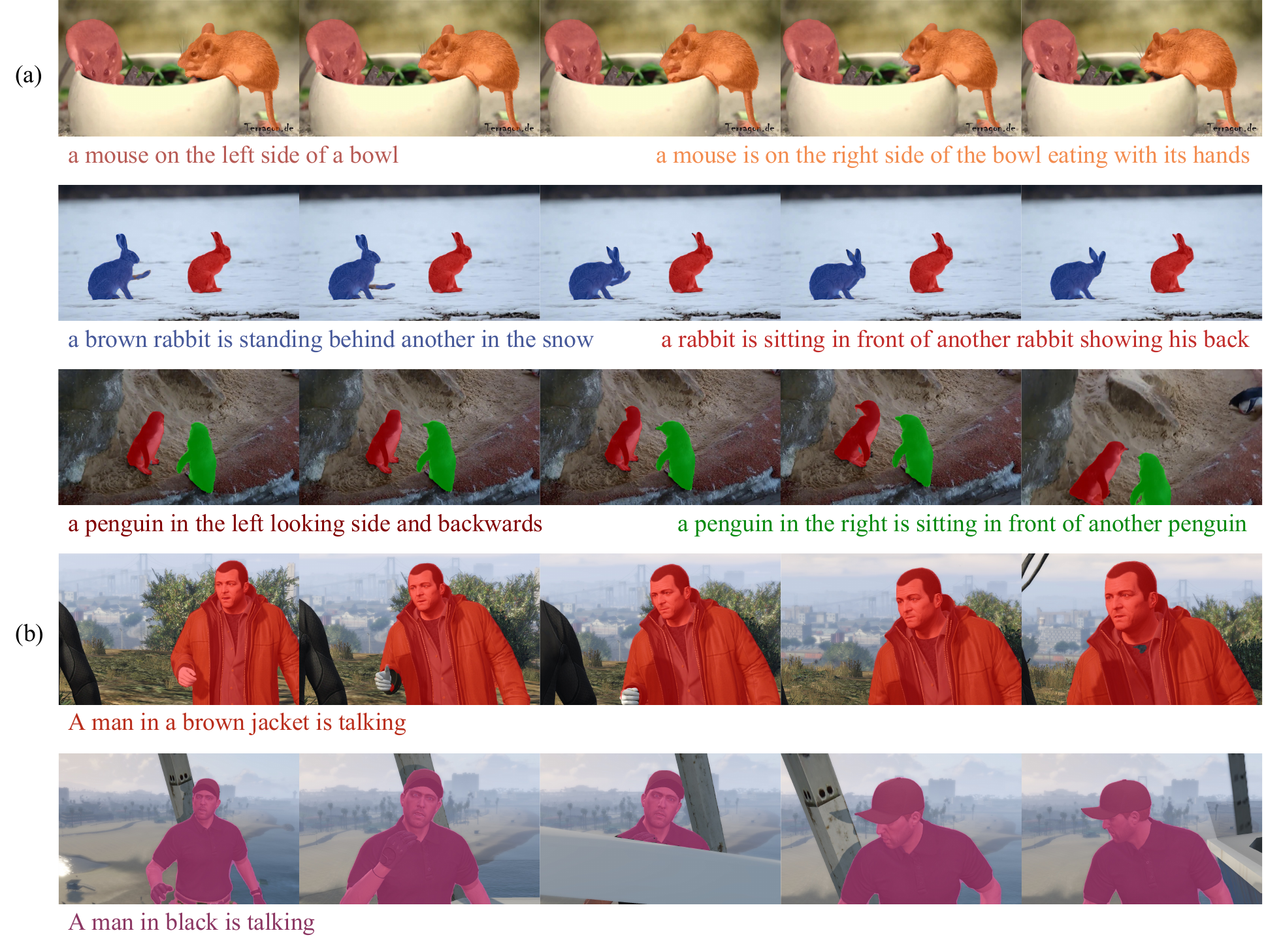}
    \vspace{1ex}
    \caption{Qualitative results on (a) Mini-Ref-YouTube-VOS and (b) Mini-Ref-SAIL-VOS.}
    \label{fig:figure8}
    \end{figure*}
    \subsection{Qualitative Results}\label{ssec:qualitative}

    The qualitative results of our model on the Mini-Ref-YouTube-VOS dataset are illustrated in Figure \ref{fig:figure8}. The figure vividly demonstrates our proposed model's adeptness in accurately segmenting referred objects across various challenging scenarios. These scenarios encompass lighting variations, object motion, partial occlusions, and more, yet the model consistently exhibits exceptional performance. Furthermore, we present qualitative results on the Mini-Ref-SAIL-VOS dataset, further affirming the model's generalization capability. Even when faced with samples from different scenes, the model delivers high-quality segmentation results. These findings underscore the effectiveness and robustness of our proposed method in addressing a myriad of real-world scenarios. Overall, through these qualitative results, we showcase the outstanding performance of our model in the task of single-object referring video object segmentation under limited sample conditions. \par

\begin{figure*}[ht]
    \centering
    \includegraphics[width=\linewidth]{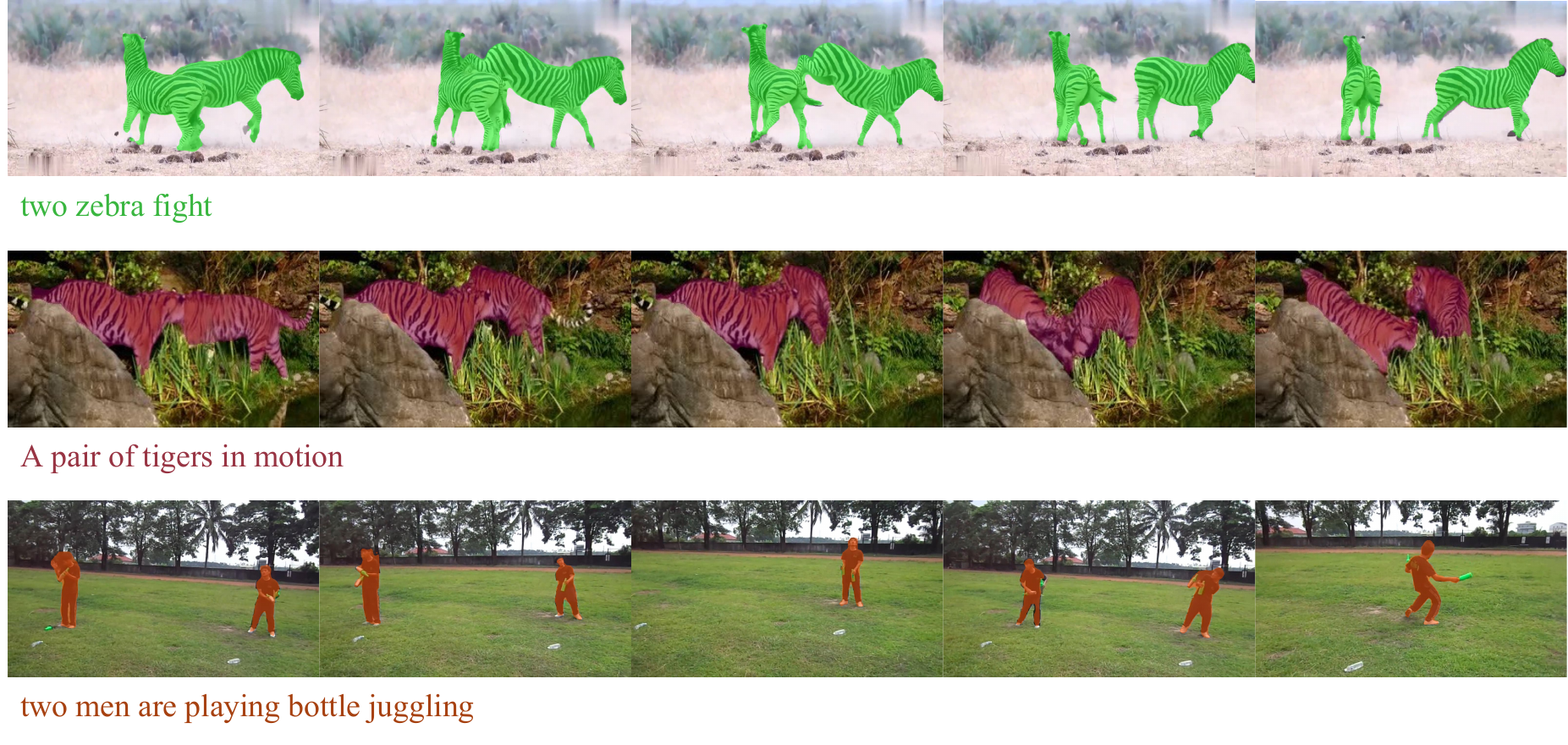}
    \vspace{0.4ex}
    \caption{Qualitative results on Mini-MeViS.}
    \label{fig:figure9}
    \end{figure*}

    The qualitative results of our model on the Mini-MeViS dataset are depicted in Figure \ref{fig:figure9}. The figure showcases the model's ability to accurately segment referred objects in multi-object scenarios across different video segments. Upon observing the examples in the figure, it's apparent that even in the presence of multi-objects, occlusions, and overlaps, our model proficiently segments each object. This indicates the model's strong robustness and generalization capability, enabling it to effectively tackle the complexities and variations present in real-world scenes. These qualitative results further validate the effectiveness and reliability of our proposed method in addressing the task of multi-object referring video object segmentation.

\section{Conclusions}
 In this work, we propose CMA with ISM structure to learn multi-modal affinity across a few samples for segmenting single-object and multi-object videos, which we generalize as the FS-RVOS and FS-RVMOS problems. We first focus on the FS-RVOS task and validate our model on newly constructed datasets - Mini-Ref-YouTube-VOS and Mini-Ref-SAIL-VOS and achieve state-of-the-art performance. Furthermore, to accommodate real-world multi-object scenarios, we extend our solution to FS-RVMOS and propose the ISM module, attached to CMA, tailored to leverage information from multiple trajectories for segmenting multi-objects. Additionally, we construct the first FS-RVMOS dataset - Mini-MeViS, and conduct extensive experiments and comparative analysis on it. Experimental results on the Mini-MeViS dataset demonstrate the superior potential of our method, achieving significant performance improvements. We hope this work inspires future research in FS-RVOS and FS-RVMOS and contributes to advancements in the field.


\section*{Acknowledgments}
{This work is partly supported by the National Natural Science Foundation of China under Grant Nos. 61971004, U21A20470, 62172136, 62122035,and 62206006.}

\section*{Data Availability Statement}
{All the benchmark datasets supporting the findings of this study are available from the corresponding author on reasonable request. Additionally, some data sets, namely the Mini-Ref-YouTube-VOS data set and the Mini-Ref-SAIL-VOS data set, are also publicly available at \url{https://github.com/ hengliusky/Few_shot_RVOS}.}



\bibliography{sn-bibliography}

\end{document}